\documentclass[a4paper,fleqn]{cas-dc}

\usepackage[numbers]{natbib}
\usepackage[dvipsnames]{xcolor}

\def\tsc#1{\csdef{#1}{\textsc{\lowercase{#1}}\xspace}}
\tsc{WGM}
\tsc{QE}

\begin{document}
\let\WriteBookmarks\relax
\def\floatpagepagefraction{1}
\def\textpagefraction{.001}

\shorttitle{REB: Reducing Biases in Representation for Industrial Anomaly Detection}    


\title [mode = title]{REB: Reducing Biases in Representation for Industrial Anomaly Detection}  



%

\author[1,2]{Shuai LYU}
\ead{shuai.lyu@connect.polyu.hk}

\author[1,2]{Dongmei Mo}
\ead{dongmei.mo@connect.polyu.hk}

\author[1,2]{Wai keung Wong}
\fnmark[*]
\ead{calvin.wong@polyu.edu.hk}

\affiliation[1]{organization={School of Fashion and Textiles, The Hong Kong Polytechnic University},
            city={Hong Kong SAR},
            country={China}
            }
\affiliation[2]{organization={Laboratory for Artificial Intelligence in Design},
            city={Hong Kong SAR},
            country={China}
}

\cortext[1]{Corresponding author}


\begin{abstract}
Existing representation-based methods usually conduct industrial anomaly detection in two stages: obtain feature representations with a pre-trained model and perform distance measures for anomaly detection. Among them, K-nearest neighbor (KNN) retrieval-based anomaly detection methods show promising results.
However, the features are not fully exploited as these methods ignore domain bias of pre-trained models and the difference of local density in feature space, which limits the detection performance. In this paper, we propose Reducing Biases (REB) in representation by considering the domain bias and building a self-supervised learning task for better domain adaption with a defect generation strategy (DefectMaker) that ensures a strong diversity in the synthetic defects. Additionally, we propose a local-density KNN (LDKNN) to reduce the local density bias in the feature space and obtain effective anomaly detection. The proposed REB method achieves a promising result of 99.5$\%$ Im.AUROC on the widely used MVTec AD, with smaller backbone networks such as Vgg11 and Resnet18. The method also achieves an impressive 88.8\% Im.AUROC on the MVTec LOCO AD dataset and a remarkable 96.0\% on the BTAD dataset, outperforming other representation-based approaches. These results indicate the effectiveness and efficiency of REB for practical industrial applications. Code: \color{blue}{\textit{{https://github.com/ShuaiLYU/REB}}}.
\end{abstract}




\maketitle












\section{Introduction}
\begin{figure}[!t]
	\begin{center}
		\includegraphics[width=3.3in]{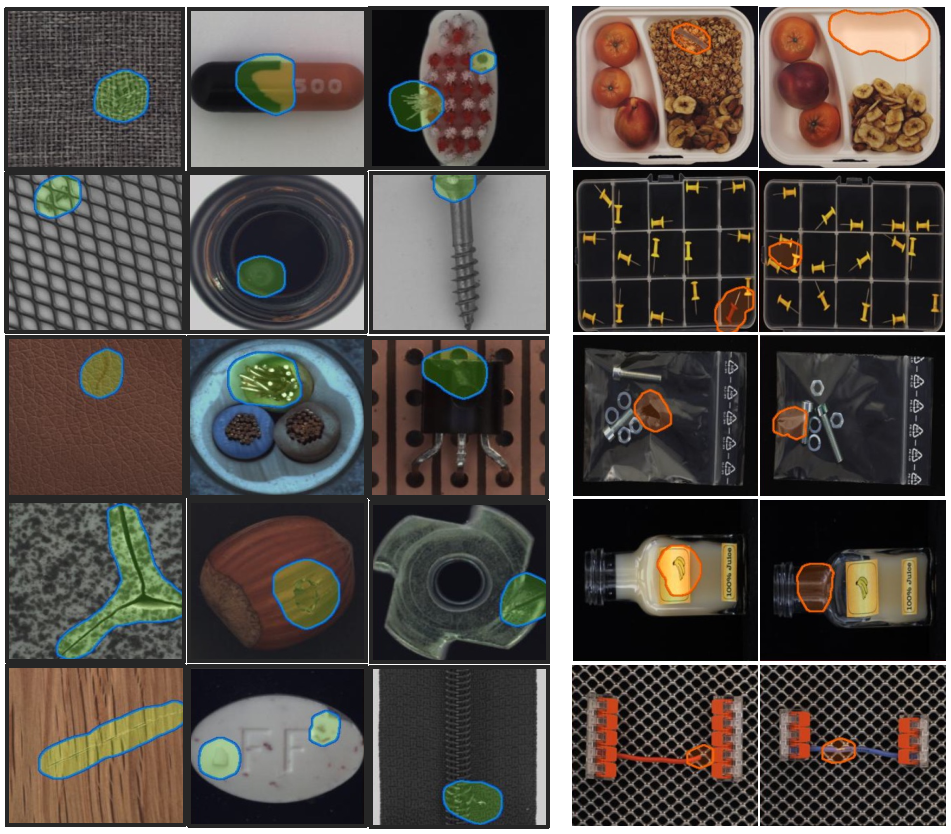}
	\end{center}
         \vspace{-0.5cm}
	\caption{
		{REB detection results on the MVTec AD (left) and MVTec LOCO AD (right) datasets. Anomalies are bounded by segmentation contours.
}
}
\label{sample_demo}
\end{figure}

Inspired by human cognition, automatic anomaly detection~\cite{pang2021deep} aims to recognize unusual patterns or destroyed structures in instances trained predominantly with normal instances. This technique has been widely applied in many fields, such as industrial product surface inspection~\cite{tao2022deep}, video analysis~\cite{SurveillanceSurvey}, and medical diagnosis~\cite{MedicalADUncertainty, UnsupervisedOCTImage}.
This work specifically targets on the industrial anomaly detection problem, i.e., detecting anomalies in industrial images. Illustrative examples from MVTec AD~\cite{bergmann2019mvtec} and MVTec LOCO AD~\cite{bergmann2022beyond}, along with detection results, are shown in Fig. \ref{sample_demo}. Based on the nature of the problem, acquiring a significant volume of anomalous data for model training is challenging. To circumvent this problem, existing works conduct unsupervised anomaly detection via auto-encoding models, Generative Adversarial Networks (GANs), and representation-based methods.
 Representation-based methods for anomaly detection consist of a feature extractor and a distribution estimator. The feature extractor extracts features from normal images or image patches with pre-trained Deep Learning (DL) models. Then the distribution estimator models the normal feature representation and measures the distance between an image's feature under detection and itself as an anomaly score. Representation-based anomaly detection has gained a lot of popularity because it is simple and effective to use pre-trained weights, which benefits from the rapid development of computer vision. Many works \cite{9479740,roth2022towards,rippel2021modeling,defard2021padim,cohen2020sub}
 just use pre-trained models to extract representations and perform distribution estimation on them for anomaly detection.
However, industrial anomaly detection has unique characteristics. Unlike natural-scene image object detection, which includes a wide variety of abstract and complex semantic information,
anomalies in industrial images often occupy a small area of the images and exhibit large irregularity, which leads to a domain bias between natural-scene image detection and industrial anomaly detection. The domain bias limits the detection performance of methods using pre-trained models. Many works~\cite{li2021cutpaste,schluter2022natural,zavrtanik2021draem,tayeh2020distance,li2020superpixel,tan2020detecting,simplenet} have explored learning a better feature representation using self-supervised learning for anomaly detection. 
In these methods, some works~\cite{li2021cutpaste,zavrtanik2021draem,tayeh2020distance,li2020superpixel,tan2020detecting} attempt to imitate natural defects for better representation learning.
However, the existing methods often fail to capture natural defects' complexity and variability fully. Although these methods attempt to generate a range of defects with different sizes and locations, the defects are often sampled from a fixed-shape distribution. Thus DL models can easily overfit and then obtain unsatisfactory performance. This work proposed a new defect generation strategy (DefectMaker) that generates a wide variety of synthetic defects for better representation learning.

On the other hand, industrial images may contain complex information, and there may be significant differences between image patches (intra-image distribution bias). Taking the MVTec AD dataset as an example, the five texture categories have smaller intra-image distribution biases, while the ten object categories contain larger intra-image distribution biases, as depicted in Fig. \ref{sample_demo}. This also results in a significant density bias in the patch-level features extracted from a CNN.
Hence, designing an appropriate distribution estimator model is a crucial step for representation-based methods, which model the patch-level features for anomaly localization and detection. A lot of work has been proposed to develop different solutions for addressing this problem. Previously, some works adopted to Gaussian density estimation (GDE)~\cite{rippel2021modeling,defard2021padim,li2021cutpaste,zheng2022focus} model the CNN features for anomaly detection. The vanilla GDE method is considered too simple and can only handle features with a simple distribution. Gaussian mixture models can theoretically handle complex distributions, but they are challenging to train for unsupervised anomaly detection problems. Some works adopted normalizing flows to map the image feature into a Gaussian distribution. Some works~\cite{roth2022towards,cohen2020sub,bergman2020deep} adopted the KNN search for feature matching and measured the distance as an anomaly score. KNN-based methods, recognized for their simplicity and efficacy, have shown promising results in anomaly detection. However, existing industrial anomaly detection works only use the vanilla KNN and totally overlook the local density bias in feature space. In other fields, existing KNN variant methods, such as Kth-NN~\cite{byers1998nearest}, LOF~\cite{breunig2000lof}, LDOF~\cite{zhang2009new}, consider the local distribution of the feature space but show little improvement in industrial anomaly detection in our experiments. In this work, we follow the KNN school and propose a LDKNN model better to consider the density bias in industrial anomaly detection. In our experiments, our LDKNN defeats vanilla KNN and other KNN variants and shows state-of-the-art results.\\
%
Overall, to address the dual challenges identified earlier, we propose REB for industrial anomaly detection, comprising two innovative modules: DefectMaker and LDKNN.
DefectMaker defines a defect as fusing a shape and a fill and using various shapes and fills to generate diverse synthetic defects. In addition, DefectMaker first introduces the Bézier curve to generate diverse defect shapes and uses the saliency model to guide the defect location to generate reasonable synthetic defects. Then, the synthetic defects are used to fine-turn a pre-trained model with a self-supervised learning task to reduce the domain bias. The fine-turned model can extract better features from industrial anomaly detection tasks. LDKNN is built on the patch-level features extracted by pre-trained models and performs unsupervised anomaly detection. LDKNN considers the density bias in features with a local density model that accounts for variations in the local feature density. The local density is used to normalize distance measurements in the inference process, thereby improving the performance of anomaly detection. The proposed REB improves the detection performance with fewer parameters to achieve real-time inspection. Our major contributions are summarized as follows: \\
\indent 1) We propose the REB, a novel representation-based framework, which reduces two key biases: domain bias and local density bias in representation-based anomaly detection. REB alleviates the limitations of pre-trained CNN feature extraction and improves the accuracy of unsupervised anomaly detection in industrial images.\\
\indent 2) Considering the difficulty and scarcity of acquiring real defects and the presence of only normal samples in unsupervised anomaly detection, we propose DefectMaker by defining a synthetic defect as fusing a shape and a fill and introducing Bézier curve saliency model to generate diverse synthetic defects. These are excellent imitations of real defects and can effectively help the network learn more suitable features for anomaly detection.
\\
\indent 3) We propose LDKNN by applying a local density model that accounts for variations in the local feature density to reduce the density bias in patch-level features and better address the anomaly detection problem. It is a non-parametric method that does not rely on assumptions on the data distribution, making it adaptable to different types of data. \\
\indent 4) We evaluated the proposed REB on two real-world industrial datasets: MVTec AD, MVTec LOCO AD, and BTAD, and experimental results show that it achieves promising image-level detection. \\
The remainder of this paper is organized as follows. Section
II introduces the related works. The proposed method is detailed in Section III. Section IV presents and discusses the experimental results. The final Section V is the conclusion of this paper.

\section{Related Works}
Anomaly detection (AD) that learns from normal images is the most popular unsupervised paradigm in defect detection and can be divided into two categories: construction-based methods and representation-based methods.
\subsection{Construction-based Methods}
Image reconstruction-based method~\cite{tsai2021autoencoder,chow2020anomaly,zimmerer2019unsupervised,mei2018automatic,an2015variational} is classical in AD methods. 
It is based on the idea that the auto-encoder (AE) models the manifold in the embedding space and the reconstruction from the embedding space. The anomalies cannot be reconstructed since they are not evolved in the training process. Therefore, the difference between the image under detection and its reconstructed images represents the anomaly detection result. 
Many techniques, such as GAN~\cite{zhao2018surface,zhang2020defgan}, learnable memory bank~\cite{gong2019memorizing,park2020learning}, inpainting masked regions~\cite{li2020superpixel,yan2021learning,zavrtanik2021reconstruction}, and knowledge distillation~\cite{jiang2023masked, cao2022informative,AST} were adopted to improve the reconstruction result. Although image reconstruction-based methods succeed in several industrial scenes, it is observed that the method typically produces incorrect reconstruction results due to the lack of feature-level discriminatory information. Another limitation is that the reconstruction-based methods belong to the end-to-end learning paradigm that is not explicit enough for explanation and can only be improved by network structure designs, external constraints, and training strategies. Since this work is not based on reconstruction methods, we do not provide a more detailed introduction to them. 

\subsection{Representation based Methods}
Recently, pre-trained models on natural image datasets such as ImageNet~\cite{deng2009imagenet} or self-supervised learning tasks have become popular feature extractors and have shown satisfactory results.
Representation-based methods for anomaly detection build an embedding feature space for normal samples and then conduct the estimation or make comparisons of the features by some distribution models, such as Gaussian density estimation (GDE)~\cite{rippel2021modeling,defard2021padim,zheng2022focus} and normalizing flow (NF)~\cite{gudovskiy2022cflow,yu2021fastflow,rudolph2021same,rudolph2022fully}, and non-parametric models, like KNN~\cite{roth2022towards,cohen2020sub,bergman2020deep} and KDE~\cite{sohn2020learning,latecki2007outlier}.
KNN-based anomaly collects a normal feature dictionary from a pre-trained model and performs KNN retrieval and distance measure. 
DN2~\cite{bergman2020deep} is the first method to combine the KNN and pre-trained CNN feature extractor, showing promising image anomaly detection results. Afterward, SPADE~\cite{cohen2020sub} extracted patch-level features and performed patch-level anomaly detection and localization with KNN. 
PatchCore~\cite{roth2022towards}, following this paradigm, built a patch-level feature dictionary and named it a memory bank. Moreover, it extracts multi-hierarchy features for better feature representation and adopts a Coreset algorithm~\cite{clarkson2010coresets} to reduce inference costs. \cite{liu2023unsupervised} proposed an anomaly detection method based on self-updated memory and center clustering (SMCC) that uses a Gaussian mixture model to fit the normal features and jointly learn the backbone and update the memory bank. Similarly, the proposed REB adopts a feature memory bank for normality representation. The difference between REB and the existing memory bank-based methods is that REB applies a locally density-normalized measure for anomaly detection. It considers the density of data points in the immediate vicinity of the point rather than the density of the entire dataset. It has seen longstanding usage in fundamental clustering algorithms\cite{schubert2017dbscan,ankerst1999optics} and outlier detection~\cite{breunig2000lof,kriegel2009loop}. We assume that the memory bank (patch-level feature dictionary) extracted from normal samples is clean, in which samples with low density are less likely to be anomalies, which is opposite to outlier detection. We achieve this by using a local density model to normalize the distance measure between the feature under detection and its immediate vicinity. 


\begin{figure*}[t]
	\begin{center}
		\includegraphics[width=1\textwidth]{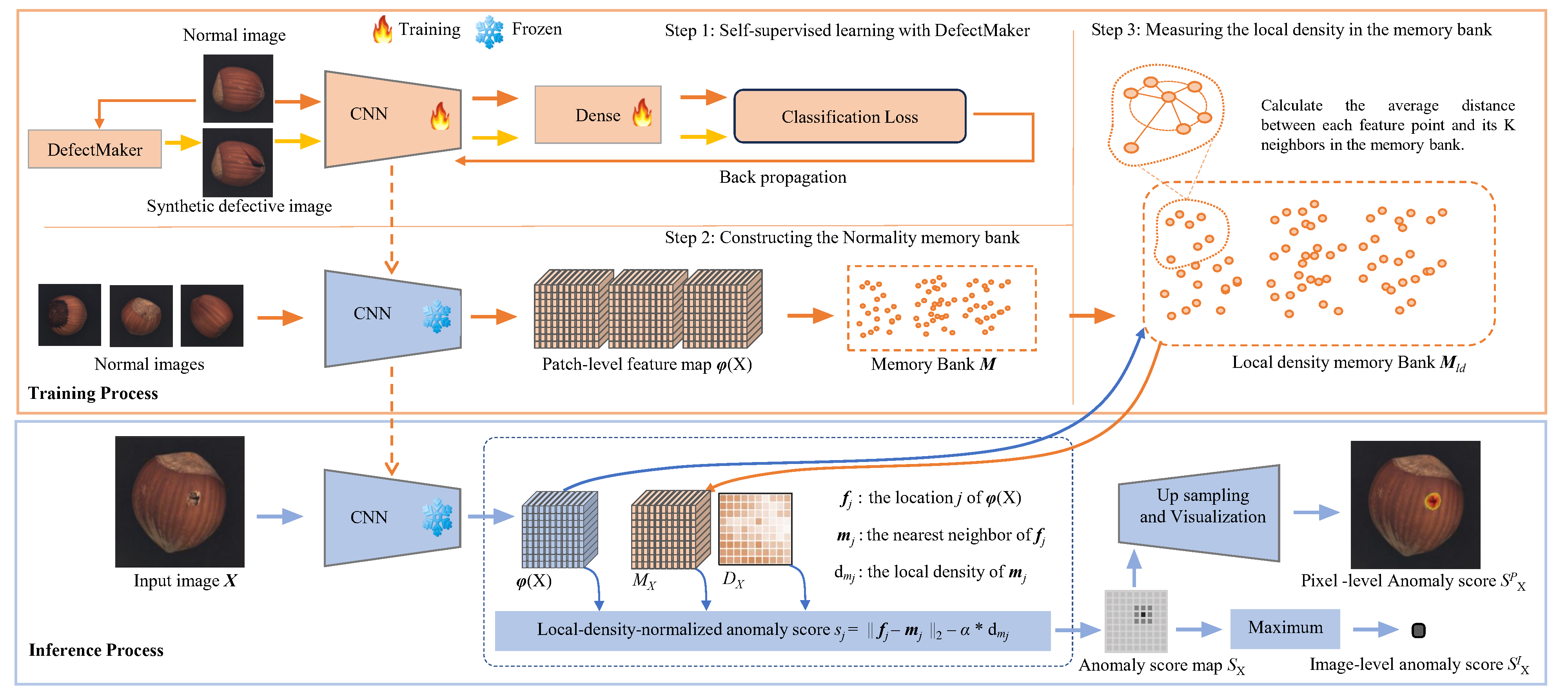}
	\end{center}
         \vspace{-0.5cm}
	\caption{
		{Overview of the framework of REB. The training process contains three steps. Step 1:  DefectMaker generates defective images, and a classification model is trained to classify between the normal and defect and then used for domain adaptation. Step 2: the pre-trained model extracts patch-level features from a group of normal images and collects them into a memory bank. Step 3: measure the local density in the memory bank. During the inference phase, CNN features of the test image are first extracted, then the nearest neighbor features for each feature are retrieved from the memory bank, and a local density-normalized distance is calculated as the anomaly score. }
  }
\label{framework}
\end{figure*}
\subsection{Synthetic Defect-based Methods}
Using pre-trained CNNs is not ideal for ignoring the domain bias between natural-scene and industrial images. Thus, many researchers proposed designing proxy tasks using self-supervised learning (SSL)~\cite{jaiswal2020survey} to reduce domain bias, which learns features from unlabeled images and then applies them to relevant visual tasks. Synthetic defect-based methods~\cite{li2021cutpaste,zavrtanik2021draem,tayeh2020distance,li2020superpixel,tan2020detecting} that use image editing to generate synthetic defect samples are the most common proxy tasks for anomaly detection. Image editing is a commonly used data augmentation technique in supervised tasks~\cite{devries2017improved,zhang2017mixup}.
The most straightforward technique~\cite{tayeh2020distance,li2020superpixel} is to randomly remove arbitrary-size regions in normal images and fill them with fixed pixel values. However, these methods generate unsatisfactory imitations of natural defects and lead to a biased model with a low generalization. There are two categories of improved techniques. The first category is increasing the training difficulty~\cite{tan2020detecting,li2021cutpaste,zavrtanik2021reconstruction,yan2021learning}. Zavrtanik, et al.~\cite{zavrtanik2021reconstruction} and Yan, et al.~\cite{yan2021learning} tried to remove a significant percentage of regions in the training process, which reduces the information redundancy in the image and increases the difficulty of the reconstruction task. So they obtained better anomaly detection performances. In addition, Tan, et al.~\cite{tan2020detecting} first proposed taking image patches sampled from another image in the same dataset as defects. Similarly, Li, et al.~\cite{li2021cutpaste} proposed an SSL method named Cutpaste, which crops the region on the original defect-free image and then pastes it onto another one to generate a new defective sample. These methods consider normal foreign patches as defects and increase the difficulty of learning tasks and achieve better feature representation.
The other category tries to generate more diverse defects. Theoretically, the more diverse the synthetic defects are, the smaller the domain bias will be. Hence, some methods practically adopted different defect distributions, sizes, brightness, and shapes, then fused them to generate more diverse defective samples \cite{zavrtanik2021draem} \cite{schluter2022natural}. For example, Zavrtanik, et al.~\cite{zavrtanik2021draem} cropped regions from various texture images of public datasets as defective fills to obtain more realistic defects. Schlüter, et al.~\cite{schluter2022natural} proposed a Poisson fusion method that addresses the apparent defect margin discontinuity caused by pasting part of one image onto another. 
Inspired by these methods, the proposed DefectMaker defines a defect as a fusing of its fill, shape, and blending into to image background to increase the training difficulty and generate more reasonable synthetic defects for better representation. In addition, DefectMaker first introduces the Bézier curve to generate diverse defect shapes and uses the saliency model to guide the defect location to generate reasonable defects.\\
\section{Method}

In this chapter, we present the REB framework through two principal processes: the training process and the inference process, as shown in Fig. \ref{framework}. The training process comprises three steps: 1) self-supervised learning with DefectMaker, 2) constructing the Normality memory bank, and 3) measuring the local density in the memory bank. In the first step, synthetic defective images are generated with the DefectMaker algorithm to reduce domain bias and facilitate the learning of industrial-targeted deep features. In the second step, the CNN features of normal images are collected and stored to construct a memory bank serving as the normality representation. In the third step, the local density of each feature point in the memory bank is measured. For the inference phase, deep features of the test images are initially extracted by the frozen CNN, followed by retrieval of the nearest neighbor features for each feature from the memory bank, calculating the anomaly score based on the local density-normalized distance.
We summarize the measuring of the local density in the memory bank process and LDKNN inference process as the LDKNN algorithm.
\subsection{Self-Supervised Learning with DefectMaker}
DefectMaker generates large amounts of synthetic defect samples and the model is trained to distinguish these synthetic images from normal ones through a proxy self-supervised learning (SSL) task.
\subsubsection{DefectMaker}
\begin{figure}[t]
	\begin{center}
		\includegraphics[width=.45\textwidth]{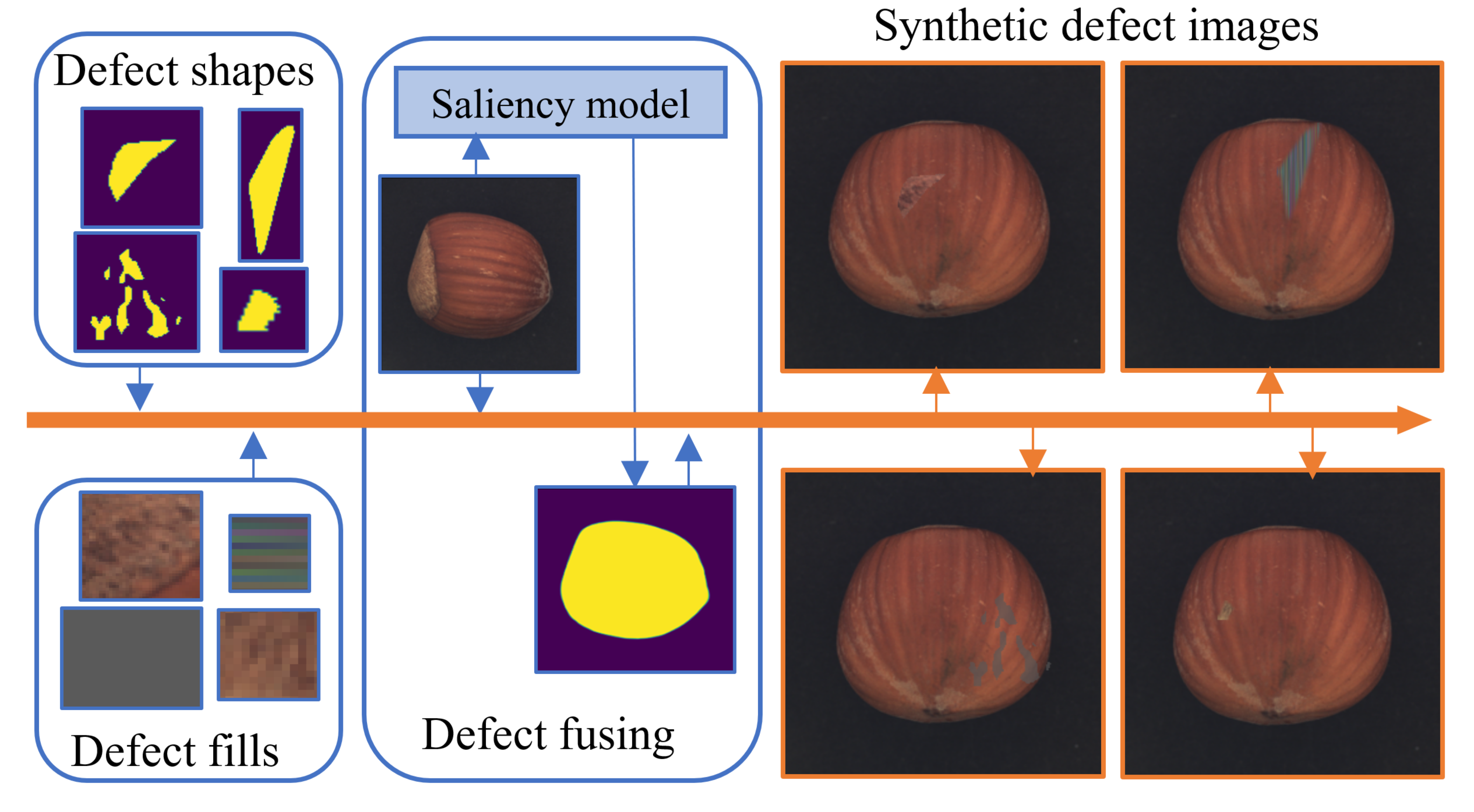}
	\end{center}
        \vspace{-0.5cm}
	\caption{\small
		{DefectMaker Pipline.}
  }
\label{fuse_model}
\end{figure}

DefectMaker is designed to generate various types of defects, including but not limited to noise, mask, distortion, and defects. To maximize the utilization of normal samples and imitate natural defects with more varieties, DefectMaker models a defect as a combination of its fill and shape with a defect-fusing strategy. As shown in Fig. \ref{fuse_model}, the DefectMaker pipeline contains three steps: 1) generate defect shape; 2) generate defect fill; 3) synthesize defect images.

\textit{Bézier shape.} Generally, the rectangles utilized in~\cite{li2021cutpaste} are overly simplistic, leading to CNNs trained with these methods exhibiting limited generalization performance. Bézier curve~\cite{farin1983algorithms} is employed to generate various shapes, aiming for enhanced simulations. A Bézier curve, a parametric curve employed in computer graphics field, simulates real-world curves using a set of discrete 'control points', rather than through a mathematical representation.
Two derivative shapes are defined: Bézier-scar and Bézier-clump shape. Scar augmentation, introduced in Cutout~\cite{devries2017improved}, utilized a long, thin rectangle to simulate tiny defects. A Bézier-scar shape is defined as a curved shape generated by the Bézier algorithm and bounded by a rectangle-scar. This approach ensures that the generated defects are more realistic, closely resembling real scratches. "Bézier-clump shape" is defined as a curve generated by the Bézier algorithm, subsequently eroded and divided into a clump of defects.\\
\textit{Defect fill.} We generate fills of two different distributions. Random noise fill refers to a type of noise that is added to digital signals or images to introduce some level of randomness or variability. It is also a popular data augmentation method in machine learning tasks. This study creates a random noise fill by controlling the properties, such as the mean and fluctuation range. Cutpaste fill~\cite{li2021cutpaste} involves obtaining the fills by cutting a region from another image from the same dataset. This technique increases the learning task's difficulty to enhance the model's discriminative ability.

\textit{Defect fusing.} DefectMaker utilized an unsupervised saliency model~\cite{li2016visual} to \textbf{identify} salient regions in images containing objects of interest, subsequently used to constrain the defect area. Saliency models involve identifying visually salient objects or distinguishing the foreground using an unsupervised paradigm in images. This study employs the EDN model~\cite{wu2022edn} as the saliency extractor. In the case of texture images, the entire image area is considered a saliency region by default. The fusion of a synthetic defect into an image is also crucial. Two fusing styles were adopted: Pasting and Blending. Pasting the targeted region directly is the most straightforward method, which can create a noticeable boundary between the defect and the background. Another method involves blending the defect with the background by using a weighted combination~\cite{zhang2017mixup}, which can yield a more seamless integration between the defect and the surrounding region. DefectMaker employs both fusing techniques simultaneously, enhancing the diversity of synthetic samples.

\subsubsection{Self-supervised learning (SSL) Task}
DefectMaker can synthesize different defects with varying sizes, shapes, textures, positions, and appearances. We define an SSL task, learning from the synthetic images, to fine-tune the pre-trained CNN models on ImageNet. In addition, defects with different DefectMaker configurations (shapes and fills) can be viewed as various categories. A multi-class classification training strategy is adopted to fully use the various types of synthetic defects. The learning objective is the popular cross-entropy function as follows:

\begin{equation}
 \begin{split}
    &L = Y*\log \left(p\left(Y\mid X_{i},\Phi\right)\right)\\
  &+(1-Y) * \log \left(1-p\left(Y\mid D M\left(X\right), \Phi\right)\right),\\
\end{split}
\end{equation}
where $X$ is a normal image, $Y$ is the label, $DM(\cdot)$ is the DefectMaker algorithm, and $\Phi$ is the CNN feature extractor.



\subsection{Constructing the Normality memory bank}
After SSL learning, the second step of the training process involves using the backbone (CNN layers) to extract patch-level features from normal images and follow~\cite{cohen2020sub} and~\cite{roth2022towards} to build a memory bank as the normality representation with the normal features. This work employs Resnet~\cite{he2016deep,zagoruyko2016wide} as the feature extractor. Let $X$ denote a normal image, and $\varphi_j (X)\in R^{C^j\times H^j\times W^j } (j\in \{1,2,3,4\}) $ is the feature map of $X$ at the $j$-th hierarchy, where $C^j$, $H^j$, and $W^j$ are the feature channel, height, and width. A commonly used strategy done by~\cite{cohen2020sub,li2021cutpaste,bergman2020deep} takes $\varphi_4$, the feature map from the last hierarchy of the Resnet18. However, this strategy may not suit industrial anomaly detection as the deeper features at the last hierarchy are often biased toward the proxy task. Anomalies may have the characteristics of small spatial size and irregularity, which are lost gradually as the network becomes deeper. This work, following~\cite{roth2022towards} , adopts the second and third hierarchies $\varphi_2 (X)$ and $\varphi_3 (X)$ for $X$. To maximize the utility of the multi-hierarchy feature maps, a feature aggregation module is introduced to aggregate the features of multiple hierarchies. Specifically, the aggregation has two steps: local spatial aggregation and multi-hierarchy aggregation. First, a local average pooling operation is applied to $\varphi_2 (X)$ and $\varphi_3 (X)$ to obtain a larger receptive field followed by upsampling $\varphi_3 (X)$ to match the resolution of $\varphi_2 (X)$. Then, $\varphi_2 (X)$ and $\varphi_3 (X)$ are aggregated on the channel dimension into an integrated feature map. $\varphi(X)=\varphi_2 (X) \oplus \varphi_3 (X) \in R^{C'\times H'\times W'} $, where $C'$, $H'$, $W'$ are equal to $C^2+C^3$, $H^2$, $W^2$. A location $\varphi(X,h,w) \in R^{C'}$ is a feature vector associated with a patch in the image. Finally, we collect the feature vectors of the training normal images into a feature dictionary (memory bank $\textbf{\textit{M}}$).

\begin{table}[t]
\centering
\small
\begin{tabular}[t]{l}
\hline
\textbf{Algorithm 1: LDKNN algorithm} \\
\hline 
\textbf{Training Process}:\\
\quad \quad \textbf{Input}: Patch-level Memory Bank $\textbf{\textit{M}}$ \\
\quad \quad $\textbf{M}_{ld}=\{empty set\} $\\
\quad \quad \textbf{For} $\textbf{m}_i \in \textbf{\textit{M}}$ do\\
\quad \quad \quad \quad retrieve the KNN $\textbf{\textit{m}}_i^1, \textbf{\textit{m}}_i^2, ..., \textbf{\textit{m}}_i^K$\\
\quad \quad \quad \quad $d_{m_i} = mean_{K}(||\textbf{\textit{m}}_i-\textbf{\textit{m}}_{ik}||_2), 1 \le k \le K$ \\
\quad \quad \quad \quad $ \textbf{\textit{M}}_{ld} \leftarrow \textbf{\textit{M}}_{ld} \cup \{(\textbf{\textit{m}}_i, d_{m_i})\} $\\
\quad \quad \textbf{Return} Memory bank with local density $\textbf{\textit{M}}_{ld}$ \\
\hline
\textbf{Inference Process}:\\
\quad \quad \textbf{Input}: Input image $X$\\
\quad \quad Patch-level anomaly score map $\textbf{\textit{S}}_{X}=\{empty set\}$\\
\quad \quad Patch-level feature map $\varphi(X) = \{\textbf{\textit{f}}_j\}, 1 \le j \le \Grave{H} \times \Grave{W}$ \\
\quad \quad \quad \quad $ (\textbf{\textit{m}}_j,d_{m_j})=\underset{(\textbf{\textit{m}}_i,d_{\textbf{\textit{m}}_i})\in \textbf{\textit{M}}_{ld}}{argmin} ||\textbf{\textit{f}}_j-\textbf{\textit{m}}_i||_2$\\
\quad \quad \quad \quad ${s_{f_j}}=||\textbf{\textit{f}}_j-\textbf{\textit{m}}_j||_2-\alpha\ast d_{m_j}$\\
\quad \quad \quad \quad $\textbf{S}_{X} \leftarrow \textbf{\textit{S}}_{X} \cup {s_{f_j}}$\\
\quad \quad Image-level anomaly score: $S_{X}^I=max(\textbf{\textit{S}}_{X})$\\
\quad \quad \textbf{Return} $\textbf{\textit{S}}_{X}^I, S_{X}$ \\
\hline
\end{tabular}
\label{LDKNN_algorithm}
\end{table}

\subsection{Local density measurement in the memory bank $\textbf{\textit{M}}$}

The vanilla KNN-based anomaly detection process involves retrieving the $K$ most similar feature points for a query feature and calculating the average Euclidean distance as the anomaly score. However, using a vanilla KNN is too ideal for assuming the patch-level feature space to be homogeneous. This assumption does not consider the local density bias stemming from the complex intra-image distribution and irregular characteristics of defects.
In the final training step, local density measurements are conducted for each feature point in the memory bank. Specifically, for each feature point $d_{m_i}$, Its $K$ nearest neighbor features (KNNs) are retrieved and the average distance between it and its KNNs is calculated as the local density by
\begin{equation}
d_{m_i}={mean}_{K}\left(||\textbf{\textit{m}}_i-\textbf{\textit{m}}_{i}^{k}||_2\right),1\le k\le K,
\end{equation}
where $m_{i}^{k}$ denotes the $k$-th neighbor of $m_i$ and
$mean_K(\cdot)$ is the mean function to calculate the average Euclidean distance between $m_i$ and the $K$ neighbors.
The distance measures the local density in the memory bank. The smaller the distance, the bigger the density. Then, a local density-based memory bank $\textbf{\textit{M}}_{ld}$ is obtained, where each feature point has its local density attribute value.

\subsection{LDKNN inference process for anomaly detection}
After the training process, two core components are obtained: a CNN feature extractor and a local density memory bank $\textbf{\textit{M}}_{ld}$ for unsupervised anomaly detection. 
In the inference process in Fig. \ref{framework}, for each testing image $X$, each testing image $X$ is first fed into the CNN feature extractor to obtain the aggregated feature map $\varphi(X)$. For each feature point $\textbf{\textit{f}}_j$ in $\varphi(X)$,1-NN is used to retrieve the most similar feature in the memory bank, and the Euclidean distance between them is determined as the anomaly score. Further, our local-density-normalized anomaly score $s_{\textbf{\textit{f}}_j}$ for patch feature $\textbf{\textit{f}}_j$ is calculated as
\begin{equation}
\label{equ_ldn}
{s_{f_j}}=||\textbf{\textit{f}}_j-\textbf{\textit{m}}_j||_2-\alpha\ast d_{m_j},
\end{equation}
where $( \textbf{\textit{m}}_j,d_{m_j})$ is the nearest neighbor of $f_j$ in the memory bank $M_{ld}$,
\begin{equation}
 (\textbf{\textit{m}}_j,d_{m_j})=\underset{(\textbf{\textit{m}}_i,d_{m_i})\in M_{ld}}{argmin} ||\textbf{\textit{f}}_j-\textbf{m}_i||_2.
\end{equation}
The anomaly score is normalized using $d_{m_j}$, with $\alpha$ serving as the local density coefficient, reflecting the degree of density bias between dense and sparse areas. Setting an appropriate value for $\alpha$ is crucial. Traversing each feature point in the feature map and calculating its anomaly score yields a patch-level anomaly score map $S_{X}$. The image-level anomaly score ${S}_{X}^I$ for an image $X$ is calculated as the maximum of the patch anomaly score map $\textbf{\textit{S}}_{X}$ as
\begin{equation}
{S}_{X}^I = max(\textbf{\textit{S}}_{X}), x_i\in X.
\end{equation}
REB is not only good at image-level anomaly detection results and is also capable of achieving pixel-level anomaly detection results
$\textbf{\textit{S}}_{X}^P$ by upsampling $\textbf{\textit{S}}_{X}$. 
We summarize the measuring of the local density in the memory bank process and local-density normalized inference process as the LDKNN algorithm, as shown in Algorithm 1. \\

\begin{table}[ht]
  \centering
  \caption{Computational Complexity Comparison between vanilla KNN and LDKNN.}
  \label{tab:complexity_comparison}
  \begin{tabular}{l|cc}
    \toprule
     & KNN & LDKNN \\
    \midrule
    Training Time Complexity & $O(1)$ & $ O(n^2)$ \\
    Inference Time Complexity & $O( n \cdot d)$ & $O(n \cdot d)$ \\
    Space Complexity & $O(n \cdot d)$ & $O(n \cdot d)$ \\
    \bottomrule
  \end{tabular}
\end{table}

\subsection{Computational complexity of LDKNN}

In this part, we analyze the complexity of LDKNN and compare it with vanilla KNN used by Patchcore, as shown in Table \ref{tab:complexity_comparison}. Vanilla KNN's training time complexity is $O(1)$, owing to the mere storage of data without any explicit learning phase. The inference time complexity is $O(n\cdot d)$, primarily due to the need for a full scan of the dataset for each query, where $n$ represents the number of data points and $d$ denotes the dimension of the feature space. The space complexity remains at $O(n \cdot d)$, due to the storage requirements for the entire dataset. LDKNN enhances the vanilla KNN approach by incorporating local density calculations for each data point during the training phase. This involves calculating the average distance between a feature point and its $K$ nearest neighbors, thus quantifying the local density. Assuming the use of a linear search, the training time complexity of LDKNN is $O(n^2)$. In the inference phase, LDKNN extends the vanilla KNN by integrating local density into anomaly scoring. Despite this additional step, the inference time complexity remains at $O(n \cdot d)$, similar to KNN, as local density adjustment is a constant time operation. The space complexity of LDKNN is equivalent to that of KNN, at $O(n \cdot d)$, accounting for the storage of the entire training set and an additional local density value for each data point. Overall, our LDKNN is highly efficient. It enhances anomaly detection performance by learning local density, without increasing the inference time complexity and space complexity.

\section{Experiments}
We conducted a series of experiments on three datasets to evaluate the effectiveness of the REB. 
In this section, we define REB as the combination of DefectMaker fine-tuned weights and LDKNN detection. The most relevant work is PatchCore which is a combination of ImageNet pre-trained weights and vanilla KNN detection. To better compare the effects of each module, we refer to the combination of DefectMaker fine-tuned weights and Vanilla KNN detection as DefectMaker. The combination of ImageNet pre-trained weights and LDKNN is referred to as LDKNN. 

\subsection{Datasets and Metrics}
\textbf{MVTec AD dataset} is a dataset for unsupervised anomaly detection with 15 categories, including ten objects and five textures. There are 5,354 industrial product images, where 3,629 normal images are for training and validation and 1,725 for testing. The training set only contains normal images, while the testing set contains images of various types of defects and normal defects. 

\textbf{MVTec LOCO AD dataset}~\cite{bergmann2022beyond} is another dataset for industrial anomaly detection. It consists of 1,772 images for training, 304 for validation, and 1,568 for testing. The training and testing sets are used in our experiment. Different from the MVTec AD dataset, it demonstrates five more complex inspection scenarios. Each possesses two difficult anomaly categories: logical and structural anomalies.

\textbf{BTAD dataset}~\cite{vtadl} is another dataset similar to MVTec AD, which is composed of RGB images representing three distinct industrial products and consists of 1,799 images for training and 741 images for testing.

\textbf{Metrics}. In the field of unsupervised anomaly detection, the evaluation of models typically involves two main aspects: anomaly detection performance and anomaly localization capability. Anomaly performance capability refers to the model's ability to recognize whether an entire image contains anomalies, demanding that the model classifies the entire image as a single unit, determining whether it is normal or contains anomalies. Anomaly localization capability, on the other hand, refers to the model's ability to identify specific anomalous pixels within an individual image, demanding that the model conduct a binary classification (anomalous or not) for each pixel. Furthermore, for a classification task, the Receiver Operating Characteristic (ROC)~\cite{ROC} curve is a graphical tool used to display a classification model's performance across all possible classification thresholds. It is drawn by comparing the True Positive Rate (TPR) and the False Positive Rate (FPR). The Area Under the ROC Curve (AUROC)~\cite{auc-roc} can be displayed by graphing the cumulative distribution function of the detection probability on the y-axis, illustrating the comprehensive classification performance of a binary classifier model at different thresholds. Based on these, Image-level AUROC (Im.AUROC) and Pixel-level AUROC (Pi.AUROC) are used to evaluate the anomaly detection performance and the localization performance, respectively.

\subsection{Training details and Hyperparameter setting}
\textbf{DefectMaker}.

We combined 3 types of defect shapes and 2 types of defect fill to create 6 distinct types of defects. In addition to the non-defect class, we defined the DefectMaker self-supervised proxy task as a 7-class classification task. We utilized the SGD optimization algorithm with an initial learning rate of 0.03, which decreased according to a cosine curve. The input image was resized to 256$\times$256. Additionally, the batch size was set to 1024, and the model was updated for 300 iterations. Considering memory limitations, we employed a training strategy that involved multiple forward passes and one backward pass on an RTX3090 GPU. \\
 \textbf{LDKNN}. Regarding the two parameters of LDKNN, $K$, and the LD coefficient, in subsequent experimental results, on three datasets, the LD coefficient is set to $1$ unless otherwise specified. On each dataset, we set the same $K$ value for each category. Due to the varying number of training images per class on the MVTec AD and MTvec LOCO AD datasets, the optimal parameter $K$ for achieving the best results may differ across different categories. It is interesting to assign a variable $K$ for each class. As shown in Table~\ref{main_table} and~\ref{mvtecloco}, we show the results with the same $K$ and variable $K$ on different classes.

\begin{table*}[b]
\begin{center}

\caption{Anomaly detection performance (Im.AUROC and Pi.AUROC) using backbones Resnet18 and WideResnet50 on MVTec AD dataset. The best results are shown in bold.}
\label{main_table}
  \resizebox{1\textwidth}{!}{
 \setlength{\tabcolsep}{0.2mm}{
\begin{tabular}{c|c||c|ccccccccccccccc|c|c|c}
\hline
 Method & Backbone & Pi.AUROC & \multicolumn{18}{c}{Im.AUROC} \\
 & & 15 Total & carpet & grid & leather & tile & wood & bottle & cable & capsule & hazelnut & metal\_nut & pill & screw & toothbrush & transistor & zipper & 5 Texture & 10 Object & 15 Total \\ \hline \hline 

 FYD~\cite{zheng2022focus} & \multirow{16}{*}{Res18} & 97.4 & 98.3 & 97.4 & 100 & 95.4 & 98.2 & 100 & 94.3 & 93.2 & 99.8 & 99 & 94.9 & 89.7 & 99.9 & 97.2 & 96.8 & 97.9 & 96.5 & 97.3 \\
 FastFlow~\cite{yu2021fastflow} & & 97.2 & - & - & - & - & - & - & - & - & - & - & - & - & - & - & - & & & 97.9 \\
 Cflow~\cite{gudovskiy2022cflow} & & \textbf{98.1} & 98.5 & 96.8 & 98.6 & 99 & 94 & 98.8 & 99.5 & 97.6 & 98.3 & 97.4 & 95.1 & 98.4 & 92.7 & 93.5 & 97.7 & 97.4 & 96.9 & 96.75 \\
 NSA~\cite{schluter2022natural} & & - & - & - & - & - & - & - & - & - & - & - & - & - & - & - & - & & & 97.2 \\
 DifferNet~\cite{rudolph2021same} & & - & 92.9 & 84 & 88 & 99.4 & 99.8 & 99 & 95.9 & 86.9 & 99.3 & 96.1 & 88.8 & 96.3 & 91.9 & 91.1 & 88.6 & 92.8 & 93.4 & 94.9 \\
 CutPaste~\cite{li2021cutpaste} & & 96.0 & 93.1 & 99.9 & 100 & 93.4 & 98.6 & 98.3 & 80.6 & 96.2 & 97.3 & 99.3 & 92.4 & 86.3 & 98.3 & 95.5 & 99.4 & 97.0 & 94.4 & 95.2 \\
 PatchCore~\cite{roth2022towards} & & 97.4 & 99.4 & 97.7 & 100.0 & 98.9 & 98.9 & 100.0 & 97.7 & 97.9 & 100.0 & 99.6 & 90.5 & 98.1 & 100.0 & 96.5 & 97.6 & 99.0 & 97.8 & 98.2 \\ 
 SMCC~\cite{liu2023unsupervised} & & 97.9 & - & - & - & - & - & - & - & - & - & - & - & - & - & - & - & & & 97.2 \\ 
 SimpleNet~\cite{simplenet} & & 95.7 & - & - & - & - & - & - & - & - & - & - & - & - & - & - & - & & & 98.3 \\ 
 RD++~\cite{rdplus} & & 97.6 & - & - & - & - & - & - & - & - & - & - & - & - & - & - & - & & & 98.6 \\ 
 
 DeSTSeg~\cite{DeSTSeg} & & 97.9 & - & - & - & - & - & - & - & - & - & - & - & - & - & - & - & & & 98.6 \\ 

 (Our) DefectMaker & & 97.7 & 98.9 & 99.7 & 100.0 & 100.0 & 99.3 & 100.0 & 99.0 & 95.1 & 99.3 & 100.0 & 97.3 & 97.5 & 100.0 & 95.2 & 100.0 & \textbf{99.6} & 98.3 & 98.7 \\
(Our) LDKNN (K=5) & & 97.9 & 99.5 & 99.2 & 100.0 & 99.8 & 99.0 & 100.0 & 99.8 & 98.8 & 100.0 & 99.8 & 95.2 & 97.4 & 99.4 & 99.3 & 99.2 & \textbf{99.5} & \textbf{98.9} & \textbf{99.1} \\
(Our) LDKNN (Variable K) & & 98.0 & 99.6 & 99.6 & 100.0 & 99.9 & 99.0 & 100.0 & 99.9 & 99.0 & 100.0 & 99.9 & 97.0 & 98.1 & 100.0 & 99.8 & 99.6 & \textbf{99.6} & \textbf{99.3} & \textbf{99.4} \\
(Our) REB (K=5) & & 98.0 & 98.9 & 99.6 & 100.0 & 100.0 & 99.4 & 100.0 & 99.5 & 97.9 & 99.4 & 100.0 & 98.7 & 98.1 & 98.9 & 99.6 & 100.0 & \textbf{99.6} & \textbf{99.2} & \textbf{99.3} \\
(Our) REB (Variable K)& & 98.0 & 99.0 & 99.7 & 100.0 & 100.0 & 99.5 & 100.0 & 99.7 & 98.3 & 99.5 & 100.0 & 99.3 & 98.3 & 100.0 & 99.7 & 100.0 & \textbf{99.6} &\textbf{ 99.5} & \textbf{99.5} \\ \hline   

 PatchCore~\cite{roth2022towards} & \multirow{9}{*}{WR50} & \textbf{98.4} & 98.4 & 99.3 & 100.0 & 99.8 & 99.2 & 100.0 & 99.5 & 98.9 & 100.0 & 100.0 & 94.6 & 97.0 & 99.7 & 99.4 & 99.7 & 99.4 & 98.9 & 99.0 \\
 SMCC~\cite{liu2023unsupervised} & & 98.3 & 100 & 99.3 & 100 & 100 & 99.6 & 100 & 96.6 & 96.6 & 100 & 99.6 & 95.5 & 91.1 & 100 & 100 & 98.5 & 99.8 & 97.8 & 98.5 \\
 SimpleNet~\cite{simplenet} & & 98.1 & 99.7 & 99.7 & 100 & 99.8 & 100 & 100 & 99.9 & 97.7 & 100 & 100 & 99.0 & 98.2 & 99.7 & 100 & 99.9 & 99.8 & 99.5 & \textbf{99.6} \\ 
 RD++~\cite{rdplus} & & 98.3 & - & - & - & - & - & - & - & - & - & - & - & - & - & - & - & & & 99.4 \\ 
 ReConPatch~\cite{reconpatch_hyun2024} & & 98.3 & - & - & - & - & - & - & - & - & - & - & - & - & - & - & - & & & 99.6 \\ 
 THFR~\cite{THFR} & & 98.2 & - & - & - & - & - & - & - & - & - & - & - & - & - & - & - & & & 99.2 \\ 

(Our) DefectMaker & & 98.1 & 99.8 & 100.0 & 100.0 & 100.0 & 99.7 & 100.0 & 99.9 & 98.0 & 99.5 & 100.0 & 98.3 & 96.0 & 99.2 & 96.9 & 100.0 & \textbf{99.9} & 98.8 & 99.1 \\
(Our) LDKNN (K=9) & & \textbf{98.4} & 98.1 & 99.2 & 100.0 & 99.9 & 99.5 & 100.0 & 100.0 & 99.8 & 100.0 & 100.0 & 97.5 & 94.6 & 97.8 & 100.0 & 99.9 & 99.3 & 99.0 & 99.1 \\
(Our) LDKNN (Variable K) & & \textbf{98.4} & 98.4 & 99.3 & 100.0 & 100.0 & 99.5 & 100.0 & 100.0 & 100.0 & 100.0 & 100.0 & 99.3 & 97.0 & 99.7 & 100.0 & 100.0 & 99.4 & \textbf{99.6} & 99.5 \\
(Our) REB (K=9) & & 98.3 & 99.6 & 100.0 & 100.0 & 100.0 & 99.7 & 100.0 & 100.0 & 99.3 & 99.7 & 100.0 & 99.0 & 94.7 & 96.1 & 99.9 & 100.0 & \textbf{99.9} & 98.9 & 99.2 \\
(Our) REB (Variable K) & & \textbf{98.4} & 99.8 & 100.0 & 100.0 & 100.0 & 99.7 & 100.0 & 100.0 & 99.3 & 99.7 & 100.0 & 99.5 & 96.0 & 99.2 & 100.0 & 100.0 & \textbf{99.9} & 99.4 & 99.5 \\ \hline

\end{tabular}
}
}
 \end{center}

\end{table*}

\begin{table*}[b]
 \begin{center}
 \caption{ Anomaly detection performance Im.AUROC and Pi.AUROC on MVTec LOCO AD datasets. The best results are shown in bold.}
 \label{mvtecloco}

\resizebox{1\textwidth}{!}{
\begin{tabular}{c|c||c|ccccc|c|c|c}
\hline
Method & Backbone & Pi.AUROC & \multicolumn{8}{c}{Im.AUROC} \\
& & Total & \multicolumn{1}{l}{Breakfast Box} & \multicolumn{1}{l}{Screw Bag} & \multicolumn{1}{l}{Pushpins} & \multicolumn{1}{l}{Splicing Connectors} & Juice Bottle & structural & logical & Total\\ \hline \hline 
(Baseline) VAE~\cite{an2015variational} & - & & - & - & - & - & - & 54.8 & 53.8 & 54.3 \\
 (Baseline) AE~\cite{tsai2021autoencoder} & - & & - & - & - & - & - & 56.6 & 58.1 & 57.3 \\
 (Baseline) VM~\cite{steger2001similarity} & - & & - & - & - & - & - & 58.9 & 56.5 & 57.7 \\
 (Baseline) f-AnoGAN~\cite{schlegl2017unsupervised} & - & & - & - & - & - & - & 62.7 & 65.8 & 64.2 \\
 (Baseline) MNAD~\cite{park2020learning} & - & & - & - & - & - & - & 70.2 & 60.0 & 65.1 \\
 (Baseline) SPADE~\cite{cohen2020sub} & - & & - & - & - & - & - & 66.8 & 70.9 & 68.9 \\
 (Baseline) ST~\cite{bergmann2020uninformed} & - & - & - & - & - & - & - & 88.3 & 66.4 & 77.3 \\
 GCAD~\cite{bergmann2022beyond} & - & & - & - & - & - & - & 80.6 & 86.0 & 83.3 \\

PatchCore~\cite{roth2022towards} & Res18 & 72.4 & 71.8 & 93.6 & 68.1 & 65.0 & 77.3 & 80.5 & 71.7 & 76.1\\ 
PatchCore~\cite{roth2022towards} & WR50 & 74.3 & 74.1 & 93.4 & 69.2 & 66.4 & 82.2 & 83.6 & 72.9 & 78.3\\ 
PatchCore~\cite{roth2022towards} & WR101 & \textbf{76.6} & 80.6 & 95.5 & 71.0 & 68.5 & 84.7 & 85.3 & 76.5 & 80.9 \\ 
 THFR~\cite{THFR} & WR50 & & - & - & - & - & - &- & - & 86.0 \\
 SINBAD~\cite{SINBAD} & WR50 & - & 77.5 & 92.2 & 71.2 & 66.9 & 80.6 & 84.7 & \textbf{88.9} & 86.8 \\ \hline \hline 

(Our) DefectMaker & \multirow{5}{*}{Res18} & 70.8 & 75.4 & 91.8 & 72.6 & 62.4 & 80.1 & 83.9 & 71.1 & 77.5\\
(Our) LDKNN (K=45) & & 68.5 & 87.9 & 96.2 & 83.4 & 72.4 & 82.8 & \textbf{90.1} & 80.7 & 85.4 \\
(Our) LDKNN (Variable K) & & 70.28 & 88.2 & 97.1 & 84.0 & 72.4 & 84.2 & \textbf{90.9} & 81.3 & 86.1 \\
(Our) REB (K=45)& & 67.7 & 89.7 & 96.9 & 78.1 & 75.5 & 84.0 & \textbf{91.3} & 80.0 & 85.7\\ 
(Our) REB (Variable K) & & 68.4 & 89.7 & 97.3 & 78.2 & 75.6 & 84.3 & \textbf{91.4} & 80.3 & 85.9 \\ \hline 

 (Our) DefectMaker & \multirow{5}{*}{WR50} &72.8 & 77.5 & 92.2 & 71.2 & 66.9 & 80.6 & 86.9 & 71.1 & 79.0 \\
 (Our) LDKNN (K=45) & & 70.0 & 90.6 & 96.9 & 87.0 & 71.1 & 88.7 & \textbf{92.6} & 83.0 & \textbf{87.8}\\
 (Our) LDKNN (Variable K) & & 70.7 & 90.6 & 97.6 & 87.2 & 71.6 & 89.0 & \textbf{92.7} & 83.5 & \textbf{88.1} \\
 (Our) REB (K=45) & & 68.0 & 91.0 & 96.5 & 79.2 & 75.0 & 88.5 & \textbf{93.3} & 80.9 & \textbf{87.1}\\ 
 (Our) REB & & 68.52 & 91.1 & 97.3 & 79.6 & 75.3 & 88.9 & \textbf{93.2} & 81.6 & \textbf{87.4} \\ \hline 
 (Our) LDKNN (K=45) & \multirow{2}{*}{WR101} & 74.4 & 91.0 & 98.9 & 88.8 & 70.1 & 90.1 & \textbf{91.4} & 85.8 & \textbf{88.6} \\
(Our) LKDKN (Variable K) & & 73.5 & 91.1 & 98.9 & 88.9 & 70.9 & 90.2 & \textbf{91.5} & 86.1 & \textbf{88.8} \\
\hline
\end{tabular}
}
 \end{center}
\end{table*}

\subsection{Anomaly Detection on MVTec AD}
This work evaluates REB with the Resnet18 (Res18) and WideResnet50 (WR50) as backbones mainly for a fair comparison with related representation-based works. Additionally, the results indicate that we have adequately addressed the problem using smaller networks. Therefore, there was no need to employ larger networks.

Table \ref{main_table} presents the AUROC scores for various methods on the MVTec AD dataset. Notably, the last three columns provide average results across 5 Texture, 10 Object, and 15 Total categories. Additionally, the PI.AUROC scores are detailed in the third column. Utilizing Res18 as the backbone, DefectMaker alone achieves an impressive 98.7$\%$ in Im.AUROC, surpassing other representation-based methods. Specifically, DefectMaker, employing vanilla KNN for detection, shows an average enhancement of 0.5$\%$ in Im.AUROC across 15 classes compared to PatchCore alone, indicating its effectiveness in mitigating domain bias and enhancing feature representation. LDKNN, on its own, records a 99.1$\%$ Im.AUROC again outperforms other representation-based methods. When compared with PatchCore, LDKNN, which also leverages ImageNet pre-training for representation, registers an average uplift of 0.9$\%$ in Im.AUROC across 15 classes, suggesting substantial improvement in feature representation. A closer examination of texture and object classes indicates that LDKNN predominantly boosts performance in object classes with higher feature density bias. Employing REB (DefectMaker and LDKNN) yields the best results of 99.3$\%$ (K=5) and 99.5$\%$ (Variable K) in Im.AUROC. With WR50 as the backbone, PatchCore already attains a 99.1$\%$ Im.AUROC. Nevertheless, our approaches, including DefectMaker, LDKNN, or REB, still achieve marginal improvements. A comparative analysis between Res18 and WR50 backbones reveals that REB when implemented on Res18, delivers results on par with those of WR50. This indicates that our proposed REB achieves a balance between fewer parameters and faster detection speeds. Among other methods, SimpleNet~\cite{simplenet} and ReConPatch ~\cite{reconpatch_hyun2024} achieve an impressive Im.AUROC of 99.6$\%$ using the WR50 backbone. In contrast, our methods showcase exceptional performance with the Resnet18 (Res18) backbone. For instance, DefectMaker reaches 98.7$\%$ in Im.AUROC, LDKNN (K=5) scores 99.1$\%$, and REB (Variable K) attains 99.5$\%$, all outperforming SimpleNet's 98.3$\%$ on Res18. REB also surpasses the SOTA result 98.6$\%$ on Res18, set by RD++~\cite{rdplus} and DeSTSeg~\cite{DeSTSeg}. This underscores the efficiency and effectiveness of our approaches, particularly in leveraging the smaller backbone. While this study primarily concentrates on detection performance as quantified by Im.AUROC, it's noteworthy that we also observe a slight enhancement in localization performance, as indicated by Pi.AUROC in the third column.


\subsection{Anomaly Detection on MVTec LOCO AD}

The MVTec LOCO AD dataset, a notable creation by Bergmann et al.~\cite{bergmann2022beyond}, represents a significant challenge in anomaly detection. To address this, they introduced the GCAD model, which impressively achieved an 83.3\% Im.AUROC. This model underwent rigorous testing against a range of established methods, including autoencoders (AE)~\cite{tsai2021autoencoder}, variational autoencoders (VAE)~\cite{an2015variational}, the Variation Model (VM)~\cite{steger2001similarity}, the memory-guided autoencoder (MNAD)~\cite{park2020learning}, SPADE~\cite{cohen2020sub}, and the Student–Teacher (ST) model~\cite{bergmann2020uninformed}. Subsequently, SINBAD~\cite{SINBAD} was introduced, which, by employing set features to simulate the distribution of elements in each sample, adeptly identifies anomalies caused by unusual combinations of normal elements, demonstrating superior results on the MVTec LOCO AD dataset. The Template-guided Hierarchical Feature Restoration (THFR~\cite{THFR}) method optimizes anomaly detection in images by compressing and restoring hierarchical features through bottleneck compression and template-guided compensation, achieving promising results. \\
In this section, we present a comparative analysis of our approach against other methods in the field, with a specific focus on Patchcore, the method most similar to ours. Similar to our experiments on the MVTec dataset, we evaluated our approach in various scenarios, including the use of DefectMaker, LDKNN, and REB, and across different backbones: Resnet18 (Res18), WideResnet50 (WR50), and WideResnet101 (WR101). Our analysis not only considers overall performance but also delves into the specific categories of structural and logical anomalies. When using Res18 as the backbone, DefectMaker alone achieved an Im.AUROC of 77.5, along with 83.9 and 71.1 for the structural and logical classes, respectively, which is inferior to GSAD and SINBAD but outperforms Patchcore using Res18. However, DefectMaker outperforms both of them on structural class. Further analysis reveals that DefectMaker's self-supervised training effectively improves the detection performance in the structural anomaly category but leads to a decrease in performance for detecting logical anomalies since DefectMaker is designed to generate structural anomalies. Moreover, when using LDKNN alone, we achieved an average Im.ARUOC of 85.4 (K=45) and 86.1 (Variable K), approaching the performance of the SOAT method SINBAD on WR50, which reaffirms the effectiveness of LDKNN in mitigating feature density bias and demonstrates its ability to capture the intricate intra-image distribution. Compared to LDKNN, REB achieves better performance in detecting structural anomalies but worse performance in detecting logical anomalies due to the influence of DefectMaker. Moreover, we achieve a better result of 88.1 Im.AUROC surpasses all other methods when using WR50 as the backbone. We conducted experiments using WR101 and only showcased the results of LDKNN, considering the potential negative impact of DefectMaker on the detection of logical anomalies. As shown in Table \ref{mvtecloco}, a better result on WR101 with an impressive Im.AUROC score of 88.8$\%$ is obtained.

\begin{table}[ht]
\caption{Comparison of state-of-the-art models on the BTAD dataset with REB, showing (Im.AUROC, Pi.AUROC) for each class. The best results are shown in bold.}
\centering
\resizebox{0.48\textwidth}{!}{
\setlength{\tabcolsep}{4mm}{
\begin{tabular}{c|c|c|c|c}
\hline
& Class 01 & Class 02 & Class 03 & Mean \\
\hline \hline
VT-ADL~\cite{vtadl} & (97.6, \textbf{99.0}) & (71.0, 94.0) & (82.6, 77.0) & (83.7, 90.0) \\
SPADE~\cite{cohen2020sub} & (91.4, 97.3) & (71.4, 94.4) & (99.9, 99.1) & (87.6, 96.9) \\
PaDiM~\cite{defard2021padim} & (98, 97.0) & (82.0, 96.0) & (99.4, 98.0) & (93.7, 97.3) \\
FastFlow~\cite{yu2021fastflow} & (99.4, 97.1) & (82.4, 93.6) & (91.1, 98.3) & (91.0, 96.3) \\
PyramidFlow~\cite{PyramidFlow} & (\textbf{100}, 97.4) & (88.2, \textbf{97.6}) & (99.3, 98.1) & (95.8, 97.7) \\
RD++~\cite{rdplus} & (96.8, 96.2) & (\textbf{90.1}, 96.4) & (\textbf{100}, 99.7) & (95.6, 97.4) \\
PatchCore~\cite{roth2022towards} & (98, 96.9) & (81.6, 95.8) & (99.8, 99.1) & (93.1, 97.3) \\
ReconPatch~\cite{reconpatch_hyun2024} & (99.7, 96.8) & (87.7, 96.6) & (\textbf{100}, 99.0) & (95.8, \textbf{97.5}) \\ 
(Our) REB (K=57) & (99.6, 94.7) & (88.5, 95.6) & (99.8, 99.7) & (\textbf{96.0}, 97.2) \\
\hline
\end{tabular}
}
}
\label{tab:btad}
\end{table}

\subsection{Anomaly Detection on BTAD}

Table \ref{tab:btad} presents a comparative analysis of REB against several SOTA methods based on WR50 backbone on the BTAD dataset which comprised 2,540 images across three different categories of industrial products. Compared with the SOTA methods (i.e. PaDiM [14], RD++ [53], and ReConPatch [22]), REB obtains competitive performance in terms of Im.AUROC and Pi.AUROC. 
Although the SOTA methods have their strengths in particular categories, REB distinguishes itself with an impressive average Im.AUROC score of 96.0\%, underscoring its superior capability in detecting image-level anomalies. Additionally, REB shows its ability to pinpoint anomalies at the pixel level with an average Pi.AUROC of 97.2\%.

\subsection{Various Feature Representations}
In this section, we explore the performance of our method with various feature representations from pre-trained ImageNet, DRAEM, CutPaste, and DefectMaker with different defect configurations. ImageNet representation refers to directly using pre-trained weights on the ImageNet dataset. Other representations refer to fine-tuning ImageNet using self-supervised tasks to reduce domain bias. For a fair comparison, we used the same backbone (Res18) and training strategy when fine-turning, with the only difference being the method of generating synthetic defects. We evaluated these different representations on three different anomaly detection methods: LDKNN, KNN, and GDE, as shown in Table \ref{dm}. Following the convention, the GDE method utilized the features from the last layer ($\varphi_4$) of Res18, while LDKNN and KNN utilized features from shallower layers ($\varphi_2$ and $\varphi_3$). By evaluating the results of GDE and LDKNN separately, we can evaluate the performance of deeper and shallower features. Obviously, LDKNN can obtain better performance even with shallower features on the MVTec AD dataset.
CutPaste uses rectangular shapes (Rect and RectScar) as anomaly shapes and DRAEM uses Perline noise to generate anomaly shapes, while DefectMaker uses more diverse Bezier shapes. In addition to CutPaste fill, DefectMaker uses random noise fill to increase the diversity of synthetic defect images. 
By comparing the results, we found that DefectMaker produces better synthetic defects and obtains superior representations. Another phenomenon is that the performance difference of the GDE method becomes more significant on different representations, which also confirms that the deeper the network hierarchy, the bigger the domain bias. Patchcore selects a shallow feature to build a memory bank to avoid this problem.

\begin{table*}[b]
  \caption{Image-level anomaly detection performance (Im.AUROC) on different feature representations on MVTec AD.}
    \begin{center}
         \resizebox{1\textwidth}{!}{
         \setlength{\tabcolsep}{1mm}{
    \begin{tabular}{l||cccc||ccc||c|c|c}

    \hline
          Representation Strategy  & \multicolumn{4}{c||}{Defect Shape} &  \multicolumn{3}{c||}{Defect Fill} &  \multicolumn{3}{c}{Im.AUROC}  \\
        \ & Rect & RectScar & { Bézier }& Perlin  & CutPaste & Random Noise & External dataset & GDE on $\varphi_4$   & KNN on $\varphi_2 + \varphi_3$  &  LDKNN on $\varphi_2 + \varphi_3$  \\ 
\hline \hline
        ImageNet & & & & & & & & 90.3 & 98.2 & 99.1 \\
        DRAEM & & & & $\checkmark$ & & &$\checkmark$ & 91.0 & 98.6 & 98.9 \\
        CutPaste & $\checkmark$ & $\checkmark$ & & &  $\checkmark$ & &  & 94.5 & 98.2 & 98.8 \\\hline
        DefectMaker & & & $\checkmark$ & & $\checkmark$ &   & &  \textbf{97.7}  & \textbf{98.8} & \textbf{99.4} \\
        DefectMaker & & & $\checkmark$ & & & $\checkmark$  & &  \textbf{96.6}  & 98.4 &  99.1 \\
        DefectMaker & & & $\checkmark$ & & $\checkmark$ &   $\checkmark$ & &  \textbf{97.3}  & \textbf{98.7} &\textbf{99.3} \\   \hline
    \end{tabular}
    }
    }
    \end{center}
    \label{dm}
\end{table*}

\begin{table}[t]
\centering
\caption{Evaluating KNN Methods for Anomaly Detection on Im.AUROC scores. The best results are shown in bold. }
\label{knns}
 \tabcolsep=2pt
    \resizebox{0.5\textwidth}{!}{

    \begin{tabular}{*{10}{l||c||cccc||c}}
        \hline
       Dataset & Representation &   KNN & Kth-NN &  LOF & LDOF & LDKNN (Our) \\
                \hline
             \multirow{2}{*}{MVTec LOCO}    &  ImageNet     & 75.1 & 74.5 & 78.3 & 78.1 &\textbf{ 83.1 } \\
            &  DefectMaker   & 75.7 & 75.8 & 78.7 & 79.8 & \textbf{84.2} \\  \hline

             \multirow{2}{*}{MVTec}   & ImageNet     &  98.2 & 97.3 & 97.2 & 96.2 & \textbf{98.8} \\

            &DefectMaker   & 98.7 & 98.6 & 96.6 & 95.8 & \textbf{99.3}  \\
                  
                \hline
    \end{tabular}
    }
\end{table}

\subsection{Various KNN variants}
Apart from the vanilla KNN, we also compared the REB on Im.AUROC with different KNN variants, i.e., Kth-NN, LOF, and LDOF.
Different from the vanilla KNN, Kth-NN uses the distance to the k-th nearest neighbor as a measure of similarity instead of the average distance to the $K$ nearest neighbors. LOF, standing for local outlier factor, is also based on the concept of local density, where the density of a point is estimated by the distance to its $K$ nearest neighbors. LDOF, standing for local density-based outlier factor, is a variation of the LOF algorithm that uses the relative location of a data point to its neighbors to measure its outlier-ness.
For all KNN variants, the neighbor size $K$ varies in a set of $\{ 3, 5, 7, 9, 11, 13, 15\} $, and the best results are reported in Table \ref{knns}. In addition, the other hyper-parameter LD coefficient $\alpha$ is fixed to 1 for a fair comparison. As shown in Table \ref{knns}, the proposed LDKNN outperforms other KNN variants on the two datasets, no matter whether the DefectMaker self-supervised learning is used. This is because the LDKNN combines distance and density metrics for anomaly detection with an LD coefficient balancing the two. In contrast, distance-based KNN and Kth-NN overlook the complex intra-image distribution. LOF and LDOF(density-based) have uneven detection ability in different density regions, with more sensitivity in low-density areas but less effectiveness in high-density areas.

\begin{figure}[!h]
\centering
\begin{center}
\includegraphics[width=.45\textwidth]{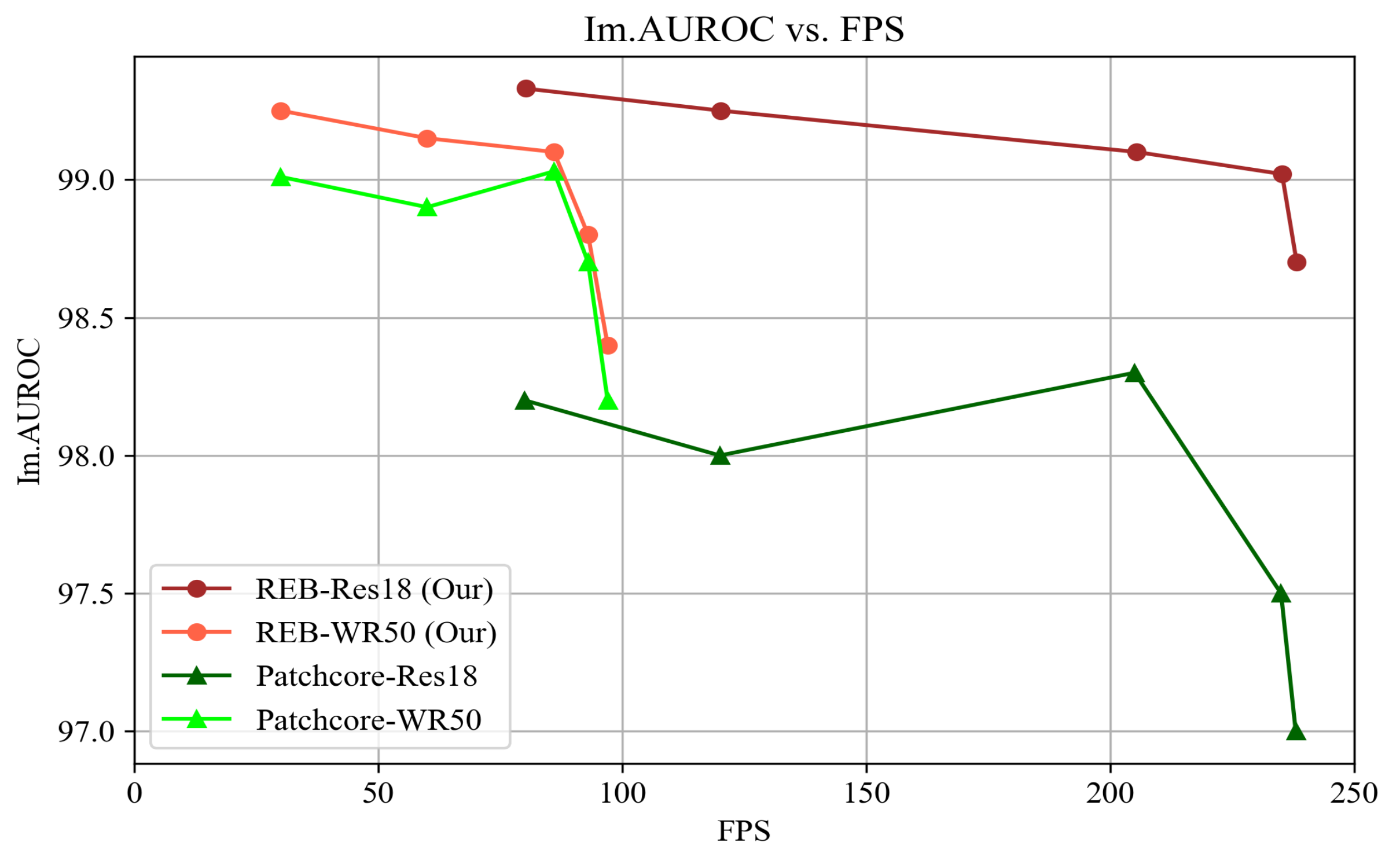}
\end{center}
\vspace{-0.5cm}
\caption{\small{Im.AUROC vs. FPS (frames per second)}
  }
\label{infer}
\end{figure}

\subsection{Coreset Algorithm and Inference Time}
Detection efficiency is crucial for industrial manufacturing, as it directly affects the production cost and efficiency. 
The inference time of KNN-based anomaly detection methods is usually positively related to the size of the backbone and memory bank. This issue has already been discussed by~\cite{cohen2020sub} and PatchCore~\cite{roth2022towards}. PatchCore adopted a Coreset algorithm~\cite{sener2017active} to find a subset to approximate the whole memory bank for lower computation cost. 
In this section, we integrate the Coreset algorithm with our REB and make a comparison with PatchCore. Similarly, we use REB-$n$\% to denote the percentage of corset. The corset proportions were set to 100\%, 50\%, 10\%, 1\%, and 0.1\% in the experiments respectively. The implementations are on the same GPU RTX3090 for a fair comparison. The Res18 and WR50 are used as the backbones of REB and PatchCore, respectively. Fig. \ref{infer} shows the relationship between FPS (frames per second) and Im.AUROC performance, where the x-axis represents the FPS, and the y-axis represents Im.AUROC. We can see that REB outperforms the PatchCore in terms of all corset proportions, especially using smaller backbone networks. When the size of the memory bank is relatively large, REB
achieves better performance because it takes into account the local distribution of the feature space more effectively. In addition, REB presents a smoother and more reliable performance curve across a range of Coreset proportions, compared with Patchcore. This observation indicates that REB is highly efficient, achieving higher accuracy without compromising inference speed. \\

\indent Focusing solely on the horizontal axis representing frames per second (FPS), the comparison also shows that, with the same backbone and coreset ratio, Patchcore and REB achieve nearly identical FPS rates. This similarity is primarily because the only difference in the model structure during the inference phase between Patchcore and REB is the employment of KNN and LDKNN, respectively. Moreover, during the inference phase, LDKNN merely adds an operation of constant value calculation. According to our statistics, the inference time difference between LDKNN and KNN is less than 0.1 ms, which is negligible. This data further confirms that LDKNN and vanilla KNN share similar inference time complexities.

\subsection{Evaluation on various  hyper-parameters $K$}
\begin{figure*}[t]
	\begin{center}
		\includegraphics[width=1\textwidth]{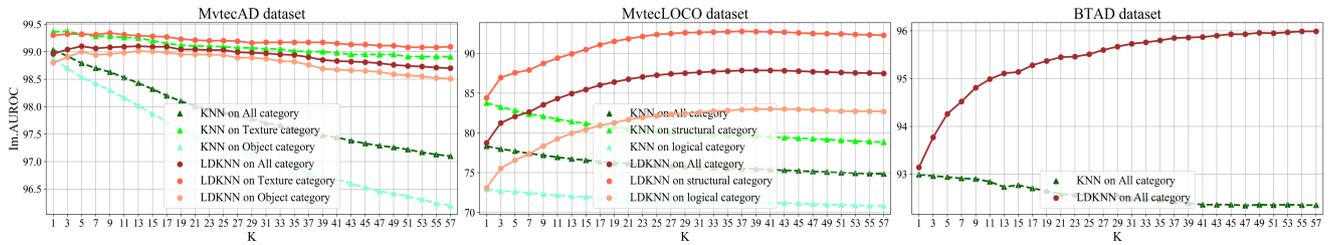}
	\end{center}
         \vspace{-0.5cm}
	\caption{Evaluating the Influence of K on Im.AUROC in LDKNN Algorithm, with KNN as the Baseline Method for Comparison.}
\label{k_exp}
\end{figure*}

In this section, the impact of varying K from 1 to 57 on the average Image-Level AUROC across three datasets: MVTec AD, MVTec LOCO AD, and BTAD, was investigated. WR50 was primarily used as the backbone. Given that KNN detection with the WR50 backbone has already shown promising results on MVTec AD, observing the impact of changes in K value on performance was deemed insufficient. Hence, on MVTec AD, we use Res18 as the backbone. Moreover, we not only analyzed the average performance across all 15 classes of the MVTec AD dataset but also delivered specific insights into the 5 texture and 10 object classes. For the MVTec LOCO dataset, we additionally differentiated the results further into structural and logical anomaly categories. In Figure ~\ref{k_exp}, LDKNN shows similar results to KNN when K is set to 1. However, LDKNN stands out as K increases, consistently improving in both anomaly detection and localization, an obvious contrast to KNN, which exhibits a notable decline in performance as K grows. This divergence is likely due to KNN's tendency to assign disproportionately low anomaly scores to high-density regions and high scores to sparse areas, thus exacerbating imbalance. Based on the results of these three experiments, LKDNN, compared to KNN, more effectively accounts for the distribution of the feature space, or in other words, it effectively reduces local density bias.

The value of K in the LDKNN is crucial in determining the number of nearest neighbors considered for calculating the local density of each feature point within the memory bank. This K value directly impacts the measurement of local densities and, consequently, the normalization of anomaly scores during the detection process. In the context of LDKNN, K affects the granularity of the local density estimation. A smaller K value focuses on the immediate vicinity of a feature point, leading to a more localized density estimation. It may also result in higher sensitivity to noise, as the local density measurement can be easily influenced by the presence of outlier neighbors. Conversely, a larger K value averages the distances over a broader set of neighbors, providing a more global perspective of the feature point's density in the memory bank, leading to more robust anomaly detection. Given that our memory bank comprises patch-level features from multiple industrial images, which may contain complex information, it often exhibits significant intra-image distribution biases between image patches. Hence, choosing a K value above 1 for the LDKNN algorithm utilizes the distribution bias to boost its effectiveness. Additionally, it has been noted that the optimal K value differs across various datasets, with performance showing a tendency to stabilize or even slightly decline as K increases. Thus, identifying the optimal K value presents a considerable challenge. This optimal value is shaped by multiple elements, such as the training set size (or memory bank size), the unique attributes of the dataset, and the specific types of anomalies being detected. In real-world applications, employing a validation set to methodically determine the optimal K value is a practical approach.

\begin{table}[t]
 \centering

 \caption{Evaluating the performance of anomaly detection with few-shot samples: Im.AUROC and Pi.AUROC metrics on MVTec AD dataset.}
 \label{fewshot}
 \tabcolsep=4pt
   \resizebox{0.45\textwidth}{!}{
\begin{tabular}{c||ccccccc}
\hline
Method \textbackslash Shots & 1 & 2 & 5 & 10 & 16 & 20 & 50 \\ \hline  
\multicolumn{8}{c}{Im.AUROC}  \\  \hline
SPADE & 71.6 &     73.4     &    75.2         &  77.5       &   78.9       &  79.6       &  81.1 \\
PaDim & 76.1  &     78.9    &    81.0         &  83.2       &   85.5        &  86.5       &  90.1 \\
DifferNet & -  &     -     &    -         &  -       &   87.3       &  -       &  - \\
Patchcore & 84.1 &     87.2     &    91.0          &  93.8       &   95.5        &  95.9       &  97.7 \\ \hline 
(Ours) DefectMaker+KNN &\textbf{ 88.5 }&    \textbf{ 91.5}     &   \textbf{ 94.6 }   &  \textbf{95.5 }  &  \textbf{ 96.0 }       & \textbf{ 96.4 }    & \textbf{ 98.4}   \\     
(Ours) DefectMaker+LDKNN & \textbf{88.6} &    \textbf{ 92.3 }    &   \textbf{ 94.9}          &  \textbf{95.7 }      &  \textbf{ 96.2 }       & \textbf{ 96.8}   & \textbf{98.6} \\
\hline
\multicolumn{8}{c}{Pi.AUROC}  \\  \hline
SPADE & 91.9  &     93.1     &    94.5         &  95.4       &   95.7        &  95.7       &  96.2  \\
PaDim & 88.2   &     90.5     &    92.5          &  93.9       &   94.8         &  95.1        &  96.3  \\
Patchcore & 92.4  &     93.3     &    94.8          &  96.1       &   \textbf{96.8  }       & \textbf{ 96.9  }     & \textbf{ 97.7 }\\ \hline
(Ours) DefectMaker+KNN &\textbf{ 93.4 }&    \textbf{ 94.8 }  & \textbf{ 95.7}    &  \textbf{96.2}    &   96.4        &  96.5     &  96.6   \\     
(Ours) DefectMaker+LDKNN & \textbf{93.4} &    \textbf{ 94.7}     &   \textbf{ 95.9}          &  \textbf{96.3}       &   96.5        &  96.6   & 96.8 \\  \hline
\end{tabular}
}
\end{table}

\subsection{Additional experiments on MVTec AD}

To explore the impact of different hyperparameters in the REB method, we present additional experiments on the MVTec AD dataset using, few-shot samples, various backbones, and hyper-parameters of LD coefficient. 

\subsubsection{Evaluation on Few-shot samples}
This section follows the Patchcore method \cite{roth2022towards} to conduct a series of experiments to evaluate the few-shot learning performance (Im.AUROC and Pi.AUROC) on the MVTec AD dataset. As shown in Table \ref{fewshot}, the proposed model is compared with four state-of-the-art anomaly detection models, under different numbers of training samples: 1, 2, 5, 10, 16, 20, and 50. Among these four baseline methods, Patchcore (ImageNet + KNN) is the best one and significantly outperforms the others. When we apply the DefectMaker algorithm to fine-tune the ImageNet-pretrained feature before KNN detection (DefectMaker+KNN), we obtain an overall improvement over Patchcore under different shots, which demonstrates its promising performance under few-shot samples. Furthermore, when we replace vanilla KNN with the proposed LKDNN, we can only observe a very small improvement. This is because LKDNN is a density-based method, which depends on a certain number of patch-level features but few-shot images form a limited patch-label feature memory bank.

\subsubsection{Evaluation on Various Backbones}
\begin{figure}[!t]
 \centering
 \includegraphics[width=0.45\textwidth]{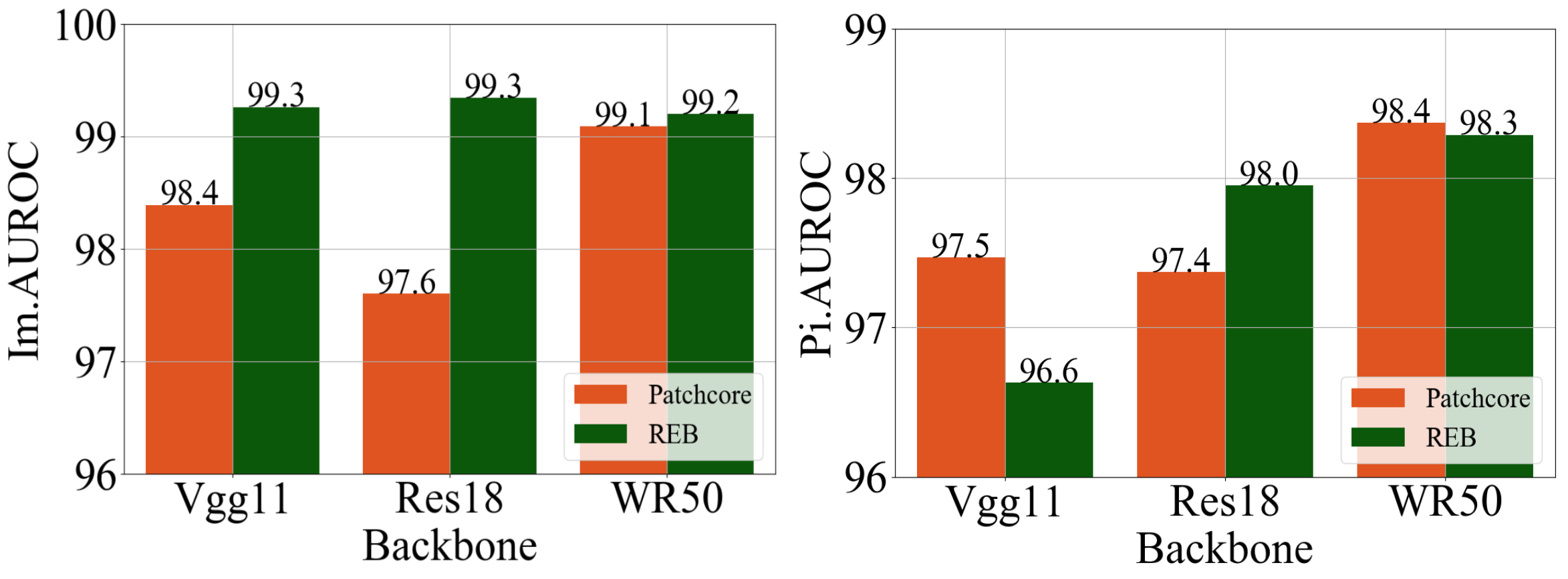}
 \centering
 \vspace{-0.2cm}
  \caption{ Im.AUROC (left) and Pi.AUROC (right) over different backbones on MVTec AD dataset.}
  \label{backbone}
\end{figure}

The performance of REB and Patchcore with different backbones (Res18, WR50 \cite{zagoruyko2016wide} and Vgg11\cite{he2016deep}) on the MVTec AD dataset is shown in Fig. \ref{backbone}. Because REB already achieves good results on the MVTec AD dataset, we did not use a larger network than WR50. REB outperforms Patchcore in anomaly detection and localization and shows outstanding performance with small backbones, enabling faster and better anomaly detection.

\begin{figure}[!t]
 \centering
 \includegraphics[width=0.49\textwidth]{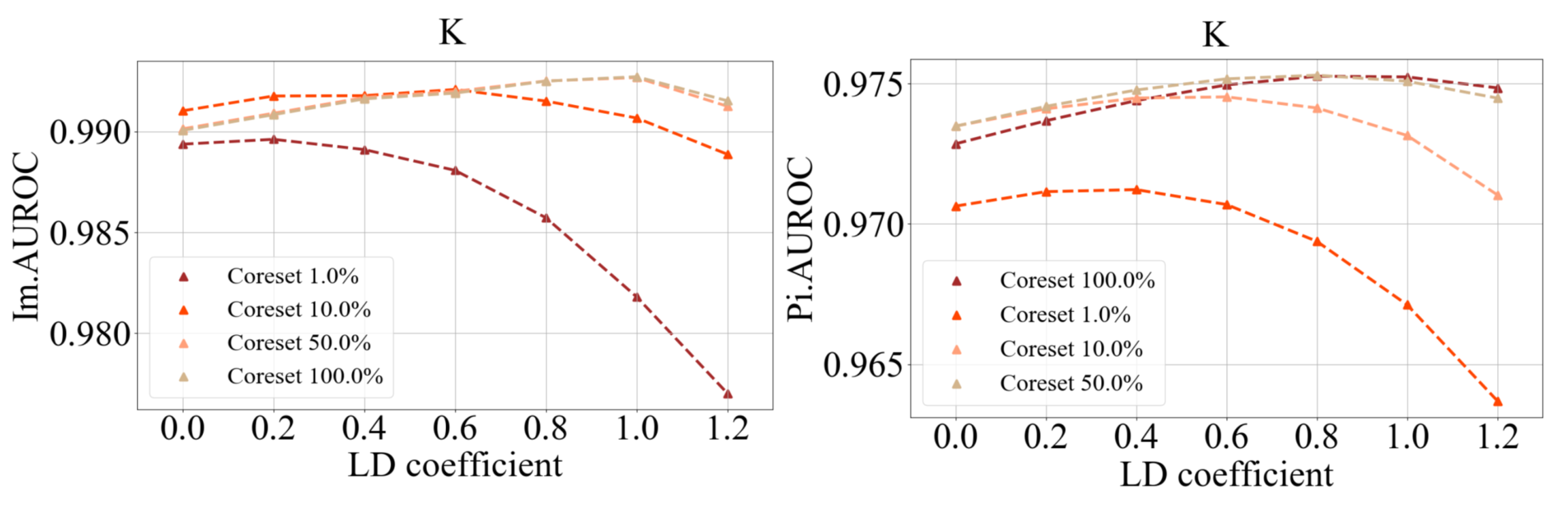}
 \centering
  \vspace{-0.5cm}
 \caption{Im.AUROC (left) and Pi.AUROC (right) over different Neighbor number $K$ and LD coefficient.}
 \label{ldknn_param}
\end{figure}

\subsubsection{Evaluation on various LD coefficient $\alpha$}
We investigate the influence of local density bias by evaluating the changes in anomaly detection performance over different LD coefficients $\alpha$ in Eq. (\ref{equ_ldn}).  Fig. \ref{ldknn_param} shows the changes in Im.AUROC and Pi.AUROC versus different values of $\alpha$, where $K$ of LDKNN is fixed to $9$ for a fair comparison. Note that when LD coefficient $\alpha=0$, LDKNN degrades to 1-NN. In addition, we also consider the relation between local density bias and memory size by involving the Coreset algorithm to down-sample the memory bank. The Coreset percentage indicates the memory bank size after down-sampling. We can see that as the $\alpha$ increases, the performance improves at the beginning and then decreases. This is because the optimums are achieved at different LD coefficients under various corset proportions. The larger the memory space size is, the larger the optimal $\alpha$ would be. The result also indicates that the local density bias degree and the memory space are positively related. 
LDKNN improves anomaly detection performance by normalizing the density bias in the feature memory bank. From another perspective, the Coreset algorithm aims to approximate the original feature space by biasing towards downsampling dense regions and preserving sparse regions, thereby reducing the local density bias across different areas in the memory bank.

\section{Conclusion}
To conclude, current approaches, such as k-nearest neighbor (KNN) retrieval-based methods using pre-trained CNN features, have limited exploitation of the feature representation in the field of industrial anomaly detection. This study proposes Reducing Biases (REB) to overcome these limitations by addressing two types of biases that exist in the pre-trained model and feature space. In the first stage, we design a DefectMaker method to generate diverse defects and a self-supervised learning task to reduce the domain bias of the pre-trained model for industrial-targeted feature representation. In the second stage, we propose a local density K-nearest neighbor (LDKNN) method to normalize the local density bias in the patch-level feature space so as to better handle complex distribution in the image and solve the anomaly detection problem. We evaluated our method on three real-world datasets: MVTec AD, MVTec LOCO AD, and BTAD datasets, and achieved significant improvements in detection performance. We demonstrate the practicality of REB by testing it on smaller CNN models, such as Vgg11 and Resnet18, achieving a 99.5\% Im.AUROC on MVTec AD and promising inference speed. Overall, our proposed REB provides an effective and practical solution for industrial anomaly detection.

Although REB has demonstrated better performance on complex data compared with SOTA methods, the result on MVTec LOCO AD does not meet the high requirement in practical applications as the Im.AUROC is less than 90\%, especially since DefectMaker is unable to simulate logical-type defects, which resulted in negative effects when evaluated on the logical class of the MVTec LOCO AD dataset. In the future, we plan to design a DefectMaker specifically targeting logical defects. The second limitation is that the two phases of REB, DefectMaker and LDKNN, are completely independent modules. We consider whether it is possible to design a self-supervised learning task and learn a memory bank instead of simply collecting features as a memory bank. In this learning process, DefectMaker can be used to learn more suitable features for the memory bank.
\\\\
\textbf{Acknowledgements}. This research is supported by Laboratory for
Artificial Intelligence in Design (Project Code: RP3-3) under InnoHK
Research Clusters, Hong Kong SAR Government.


{
\bibliographystyle{cas-model2-names}

\bibliography{egbib}

}



\end{document}